\documentclass[final]{cvpr}
\usepackage{times}
\usepackage{epsfig}
\usepackage{graphicx}
\usepackage{amsmath}
\usepackage{amssymb}
\usepackage{wrapfig}
\usepackage[ruled, linesnumbered]{algorithm2e}
\usepackage{xcolor}
\usepackage{bbm}
\usepackage{algorithmic}
\usepackage{subcaption}
\usepackage{microtype}      
\usepackage{makecell}
\usepackage{multirow}
\usepackage{mathtools}
\usepackage{algorithmic}
\usepackage{booktabs} 
\usepackage{diagbox}
\usepackage{threeparttable}

\usepackage[pagebackref=true,breaklinks=true,colorlinks,bookmarks=false]{hyperref}

\def\cvprPaperID{6387} 
\def\confYear{CVPR 2021}

\begin{document}

\title{Troubleshooting Blind Image Quality Models in the Wild}

\author{Zhihua Wang$^1$, Haotao Wang$^2$, Tianlong Chen$^2$, Zhangyang Wang$^2$, and Kede Ma$^1$\\
$^1$ City University of Hong Kong, $^2$ University of Texas at Austin\\
{\tt\small zhihua.wang@my.cityu.edu.hk, \{htwang,tianlong.chen,atlaswang\}@utexas.edu, 
kede.ma@cityu.edu.hk}
}

\maketitle
\thispagestyle{empty}

\begin{abstract}
Recently, the group maximum differentiation competition (gMAD) has been used to improve blind image quality assessment (BIQA) models, with the help of full-reference metrics. When applying this type of approach to troubleshoot ``best-performing'' BIQA models in the wild, we are faced with a practical challenge: it is highly nontrivial to obtain stronger competing models for efficient failure-spotting. Inspired by recent findings that difficult samples of deep models  may be exposed through network pruning, we construct a set of ``self-competitors," as random ensembles of pruned versions of the target model to be improved. Diverse failures can then be efficiently identified via self-gMAD competition. Next, we fine-tune both the target and its pruned variants on the human-rated gMAD set. This allows all models to learn from their respective failures, preparing themselves for the next round of self-gMAD competition. Experimental results demonstrate that our method efficiently troubleshoots BIQA models in the wild with improved generalizability.
\end{abstract}

\section{Introduction}

 \begin{figure}[t]
	\begin{center}
		\subfloat[]{\includegraphics[width=0.23\textwidth]{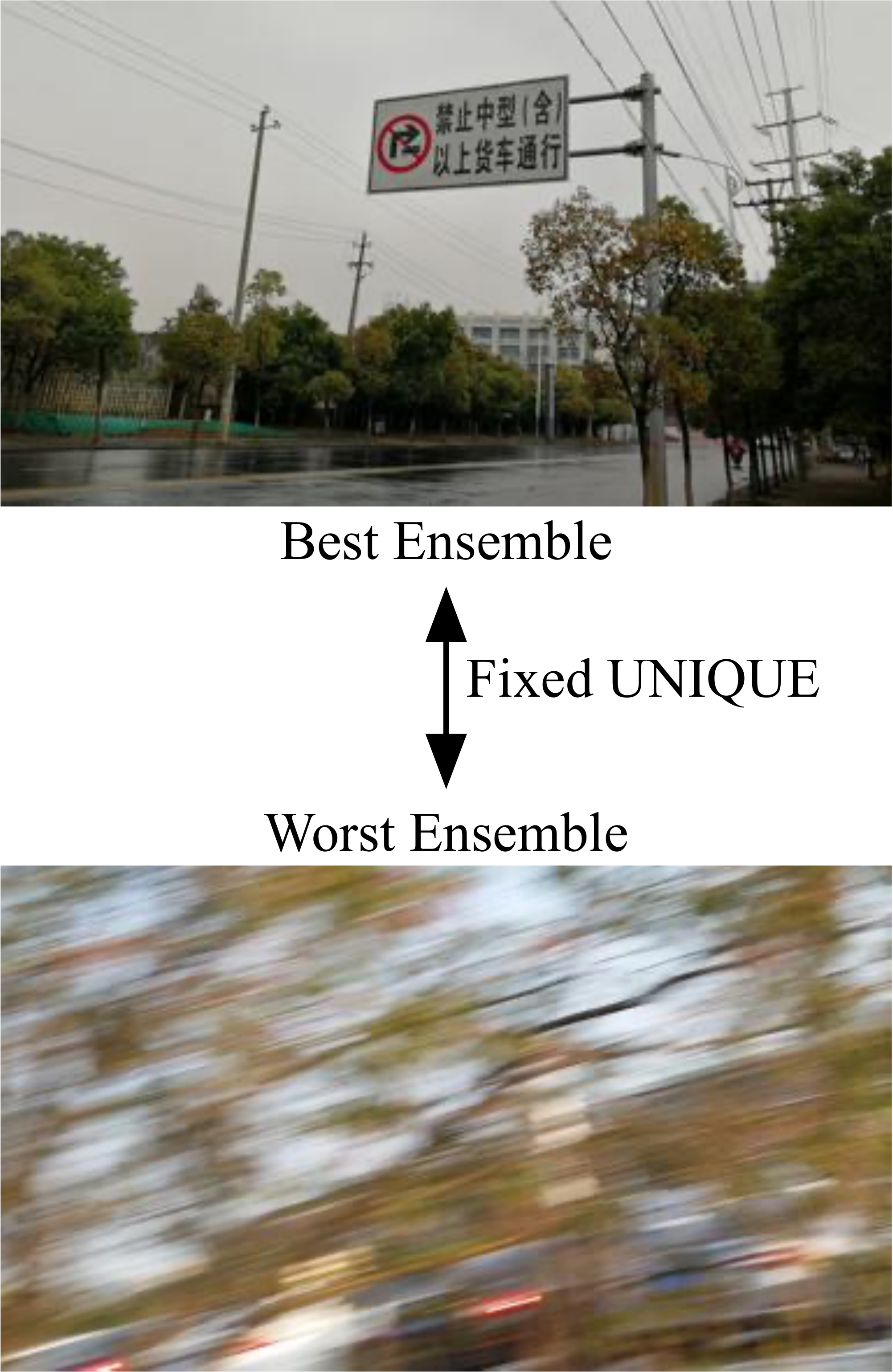}}\hskip.25em
		\subfloat[]{\includegraphics[width=0.23\textwidth]{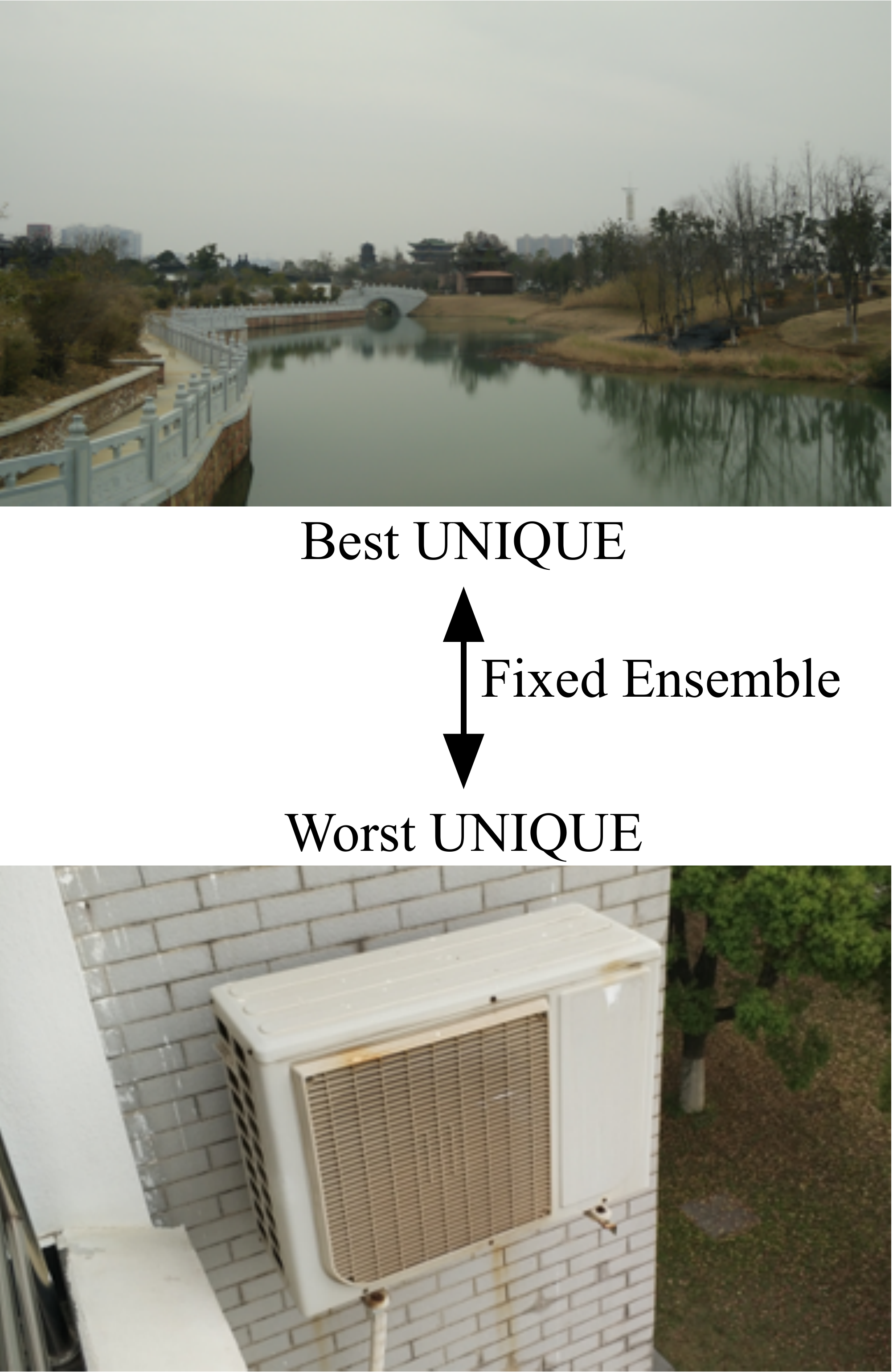}}\hskip.25em
	\end{center}
	\vspace{-.5cm}
	\caption{\textcolor{black}{Failure cases of a ``top-performing'' BIQA method - UNIQUE \cite{zhang2020uncertainty} spotted by an ensemble of its pruned versions. \textbf{(a)} Best/worst-quality images according to the ensemble, with near-identical quality reported by UNIQUE. \textbf{(b)} Best/worst-quality images according to UNIQUE with near-identical quality reported by the ensemble.}}
	\label{fig:failure}
\end{figure}
Over the years, researchers and engineers in the field of image processing and computer vision have realized the importance of blind image quality assessment (BIQA) \cite{wang2011reduced}. Numerous BIQA models \cite{bosse2017deep,ma2017end, mittal2012no, mittal2012making, ye2012unsupervised, ying2020patches} have been proposed, focusing mainly on boosting performance on existing IQA datasets of fixed sizes. However, the superior correlation numbers on closed test sets may not translate in a reliable way to generalization in the open visual world \cite{wang2008maximum,ma2018group,scheirer2008fusion,zhang2014predicting}. Therefore, computational methods for probing and improving the generalizablity of BIQA models are highly desirable.

In 2008, Wang and Simoncelli \cite{wang2008maximum} described a maximum differentiation (MAD) competition procedure to compare IQA models in the space of all possible images. Ma \etal \cite{ma2018group} proposed gMAD, a discrete instantiation of the MAD method, by restricting the search space to some specific domain of interest. Both methods are able to automatically and efficiently expose failures of a relatively \textit{weak} IQA model, by letting it compete with a set of \textit{strong} models. Wang and Ma \cite{wang2020active} took advantage of gMAD to identify the counterexamples of a BIQA model \cite{zhang2020uncertainty} using a set of stronger full-reference IQA metrics. Furthermore, they demonstrated that harnessing gMAD-selected failures significantly improves the BIQA generalizability. 

Despite demonstrated success, the progressive failure identification and model rectification pipeline proposed in \cite{wang2020active} have two drawbacks. First, it can only be applied to the synthetic distortion scenario, where full-reference IQA models are computable. For BIQA models in the wild with input images containing realistic camera distortions, it is highly nontrivial to obtain a list of stronger methods to falsify a state-of-the-art model. Second, the competing full-reference models are fixed throughout model development, rendering failure-spotting less effective as the target model becomes stronger \cite{wang2020active}.

In this paper, we present an innovative extension of the pipeline proposed in \cite{wang2020active}, to troubleshooting BIQA models in the wild. We start with a ``top-performing'' BIQA method based on deep neural networks (DNNs) as the target model. Instead of finding strong external methods, the key step in our approach is to compress the target model using network pruning techniques \cite{han2015learning, he2019filter, li2016pruning,liu2017learning}, to construct strong ``self-competitors" from the target model. The critical underlying rationale takes root in the recent finding \cite{hooker2020compressed} in image classification. The authors observed that, network pruning, which usually removes smallest-magnitude weights in a trained network, does not affect all learned classes or samples equally. Rather, it tends to disproportionally hamper the network memorization and generalization on the long-tailed and most difficult images from the training distribution. In other words, those images are not ``memorized" well by the current model, and therefore easily ``forgotten" when pruning the model. In short, network pruning can effectively spot the samples not yet well learned or represented, hence exposing the weakness of the trained model. 

Inspired by this prior wisdom \cite{hooker2020compressed}, we propose to leverage network pruning in revealing superficial ``shortcuts" in (either original or pruned) BIQA models. In order to encourage spotting \textit{diverse} failures of the target model, we create ensembles of subsets of pruned models \cite{zhou2012ensemble} to compete with the target model in gMAD \cite{ma2018group} (see Figure \ref{fig:failure}).  We then jointly fine-tune the target and all pruned variants on the combination of the human-rated gMAD images and previously trained data. This allows all competing models to learn from their respective failures, and prepare themselves for the next round of gMAD competition.


Our method is the first of its kind to troubleshoot BIQA models in the wild. The fine-tuned model shows improved aggressiveness and resistance \cite{ma2018group} in gMAD,  comparing with itself in previous rounds. In addition, we find that the images in the gMAD sets exhibit increasing transferability to falsify existing BIQA models. Our code is publicly available at \url{https://github.com/wangzhihua520/troubleshooting_BIQA}.

\section{Related Work}
\paragraph{BIQA with DNNs} Recently, there has been a surge of interest in  developing BIQA models based on DNNs. A major challenge along this direction is to constrain the large set of network parameters using a small set of human-rated images with mean opinion scores (MOSs). Kang \etal \cite{kang2014convolutional} trained a DNN with one convolution layer on $32\times 32$ patches to compensate for the lack of training data. Bosse \etal \cite{bosse2017deep} developed a DNN with more convolution layers using the same patch training strategy. Ma \etal \cite{ma2017end} leveraged the distortion identification as an auxiliary task to warm up the training. Kim \etal \cite{kim2019deep} used the error map from the Minkowski metric
to regularize the training. Ma \etal \cite{ma2019blind} took a step further, and exploited multiple full-reference metrics as noisy annotators for training DNN-based BIQA models without MOSs. These methods are mostly designed to handle synthetic distortions~\cite{sheikh2006statistical,lin2019kadid}, with limited generalizability to realistic distortions \cite{ghadiyaram2015massive,hosu2020koniq}. To meet the cross-distortion-scenario challenge, Zhang \etal \cite{zhang2018blind} bilinearly pooled two feature representations that are sensitive to synthetic and realistic distortions, respectively. Zhang \etal \cite{zhang2020uncertainty} described a simple method to train BIQA models on multiple IQA datasets. The resulting UNIQUE model is capable of assessing image quality in the laboratory and wild, and will be used as the target model to demonstrate the feasibility of the proposed method.

    
\paragraph{Network Pruning} DNNs commonly hinge on over-parameterization \cite{liu2018rethinking} and can be effectively compressed \cite{lecun1990optimal}. Network pruning \cite{han2015learning} has been an effective technique to remove redundant computation at surprisingly little sacrifice of test accuracy. 
For example, Han \etal \cite{han2015learning} proposed to prune DNNs by thresholding model \textit{weights} based on their magnitudes. Li \etal \cite{li2016pruning} pruned DNN \textit{filters} with small $\ell_1$- or $\ell_2$-norms. 
Liu \etal \cite{liu2017learning} encouraged channel sparsity by adding $\ell_1$-constraints on the batch normalization scaling parameters.
Molchanov \etal \cite{molchanov2019importance} estimated filter importance using Taylor expansion. 
He \etal \cite{he2019filter} used geometry median to select the most redundant filters. A latest review is referred to \cite{blalock2020state}.

Some researchers have started to rethink pruning beyond just an ad-hoc compression tool, and to explore its in-depth connection with DNN memorization/generalization. Frankle \etal \cite{frankle2018lottery} pioneered to show that there exist highly sparse ``critical subnetworks" from the full DNNs, that can be trained in isolation from scratch. That critical subnetwork could be effectively identified by pruning \cite{you2019drawing,frankle2019lottery}. The most relevant work is due to Hooker \etal \cite{hooker2020compressed}, who showed that pruning a trained image classifier tends to harm  its performance more on the most difficult and long-tailed training images. This implies that pruning might effectively spot samples not well learned by the current model, and provides novel insights to exposing a trained model's potential weakness. We take inspiration from \cite{hooker2020compressed}, and improve their method to troubleshoot BIQA models, by identifying and leveraging quality-discriminable images between pruned and non-pruned methods. 


\paragraph{Active Learning} The main idea of active learning is to mine the most valuable samples to label from a large unlabeled dataset \cite{cohn1994improving}. In active learning for regression,  query by committee (QBC) selects the most disagreed samples  by a committee of models \cite{seung1992query,burbidge2007active}. Expected model change maximization (EMCM)  selects samples that can cause the largest change to the current model \cite{cai2013maximizing}. Greedy sampling (GS) was originally proposed as a robust clustering method against outliers \cite{bhaskara2019greedy}. It was adapted to active learning in \cite{wu2019active}, with the goal of selecting samples that can increase the diversity of model responses. Residual active learning (RSAL) \cite{douak2011two,douak2013kernal} trains two models to fit the target outputs and the prediction residuals, respectively. The residual model is then
used to select samples with the maximum predicted residuals. The procedure of identifying gMAD images \cite{ma2018group} in this work can be seen as a form of active sampling with the criterion that the selected images have the greatest potential to falsify the target model. We will compare the error-spotting efficiency of several active learning methods in Section \ref{sec:compare_al}. 

\section{Proposed Method}
    \begin{figure}[t]
		\begin{center}
			\includegraphics[width=1.0\linewidth]{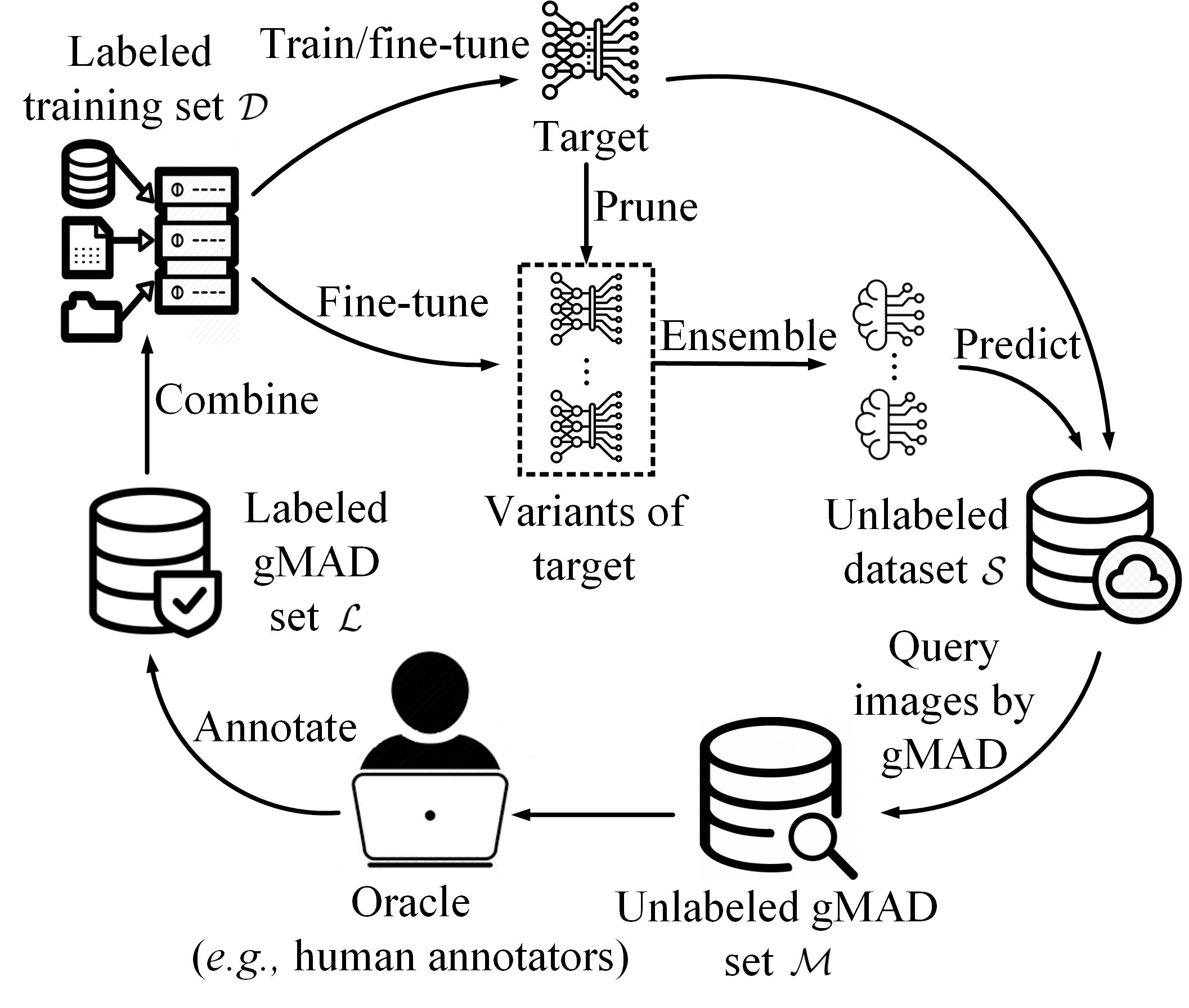}
		\end{center}
		\vspace{-.5cm}
		\caption{Diagram pf troubleshooting BIQA models in the wild. We start with a differentiable parametric target BIQA model, seek pairs of images by letting it compete with ensembles of pruned variants in gMAD \cite{ma2018group}, collect human scores for the gMAD set, fine-tune all models on the combination of the previously seen databases and the newly annotated gMAD set. The target and competing models co-evolve for the next round of troubleshooting.}
		\label{fig:active_gmad}
   \end{figure}
    We formulate the general problem of troubleshooting BIQA models in the wild as follows. We assume a strong off-the-shelf BIQA model $f$ that has been trained on the labeled set $\mathcal{D}$ with images captured in the wild. Also assumed is a large-scale unlabeled set $\mathcal{S}$, containing images with much greater scene complexities and realistic distortions. The end goals are to identify diverse failures of $f$ in $\mathcal{S}$ with a limited human labeling budget, and to leverage the exposed failures to further improve the generalizability of $f$. 
    
    The gMAD competition \cite{ma2018group} suggests to select images that optimally distinguish $f$ and a stronger competing model because those images are most likely to be its counterexamples. Then the core question is ``how to obtain a set of diverse competing models for efficient failure-spotting?'' In this paper, we create strong competing models, dubbed ``self-competitors," from the target model by network pruning \cite{hooker2020compressed}.  After obtaining the labeled gMAD set $\mathcal{L}$ through subjective testing, we jointly fine-tune the target and competing models on the combination of $\mathcal{D}$ and $\mathcal{L}$, attempting to learn from spotted failures without forgetting previously seen data \cite{li2018learning}. Figure \ref{fig:active_gmad} illustrates the proposed  diagram of troubleshooting BIQA models in the wild.
    
    \begin{figure*}[t]
	\centering
	\subfloat[$56$ / $42$ / $10$]{\includegraphics[width=0.19\textwidth]{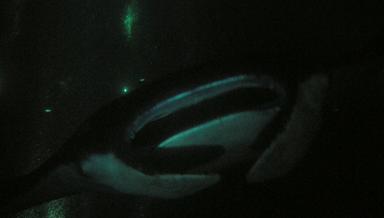}}\hskip.25em
	\subfloat[$66$ / $48$ / $17$]{\includegraphics[width=0.19\textwidth]{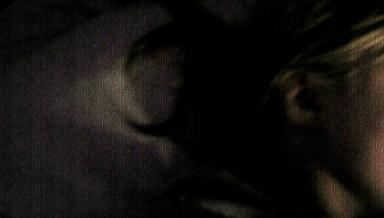}}\hskip.25em
	\subfloat[$14$ / $58$ / $37$]{\includegraphics[width=0.19\textwidth]{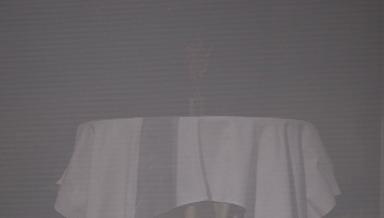}}\hskip.25em
	\subfloat[$15$ / $50$ / $38$]{\includegraphics[width=0.19\textwidth]{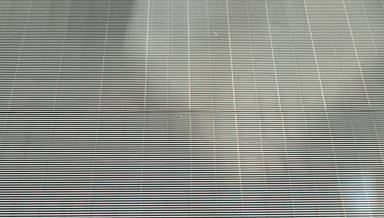}}\hskip.25em
	\subfloat[$66$ / $48$ / $30$]{\includegraphics[width=0.19\textwidth]{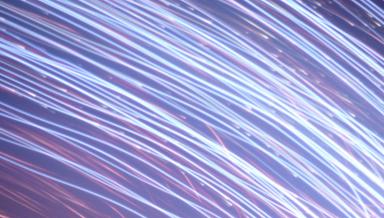}}\\
	\subfloat[$58$ / $37$ / $25$]{\includegraphics[width=0.19\textwidth]{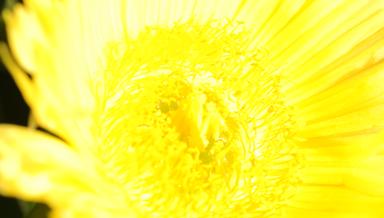}}\hskip.25em
	\subfloat[$51$ / $29$ / $20$]{\includegraphics[width=0.19\textwidth]{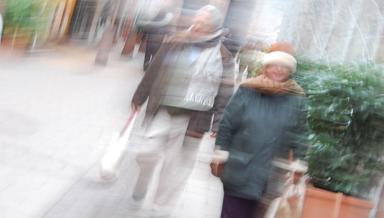}}\hskip.25em
	\subfloat[$71$ / $48$ / $42$]{\includegraphics[width=0.19\textwidth]{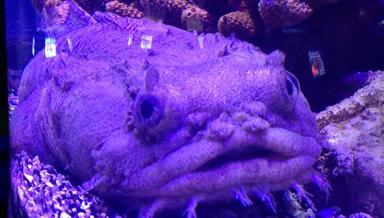}}\hskip.25em
	\subfloat[$26$ / $38$ / $69$]{\includegraphics[width=0.19\textwidth]{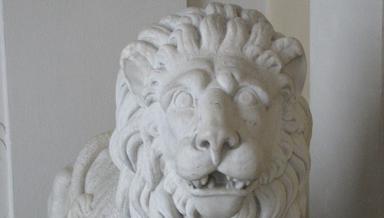}}\hskip.25em
	\subfloat[$16$ / $35$ / $55$]{\includegraphics[width=0.19\textwidth]{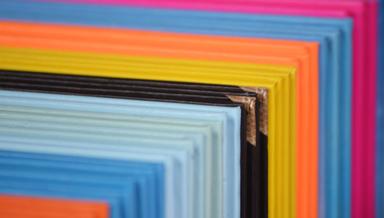}}\\
	\caption{Representative failures of the target model in gMAD. \textbf{(a)}-\textbf{(e)} Optimization results of Eq. \eqref{eq:gmad1} with $k=10$ and $n=10$. \textbf{(f)}-\textbf{(j)} Optimization results of Eq. \eqref{eq:gmad2} with $k=1$ and $n=120$, which appear much more diverse in scene content and distortion type. Below each image are scores from the target model / the competing models / humans.
	}
	\label{fig:deverse}
\end{figure*}
    
    \subsection{Self-Competitor Construction} \label{section: ensemble}
	The success of failure-spotting of the target model by gMAD \cite{ma2018group} depends  on the \textit{strength} of the competing models. Here we resort to network pruning for  competing model construction due to two reasons. First, DNN-based models are highly overparameterized. Therefore, the performance drop of pruned models on test sets is often insubstantial, and can be easily recovered  if fine-tuning is allowed. Second, Hooker \etal \cite{hooker2020compressed} showed in the context of image classification that images that differentiate between the original and pruned classifiers are the least-memorized and weakest-learned ones by the original models. They appear the most challenging for both models, and sometimes even for humans to classify. In BIQA terms, these are likely to be selected by gMAD as the most informative images to falsify both the target and competing models. 
	
	Specifically, we first generate a list of pruned models $\{h_j\}_{j=1}^m$ from the target model $f$. Thanks to the prosperity of the network pruning field, we are able to leverage a diverse set of state-of-the-art network pruning techniques \cite{han2015learning, he2019filter, li2016pruning, liu2017learning, molchanov2019importance}, with different hyperparameter settings, to encourage diversity among pruned models. Furthermore, as ensemble models have been shown to achieve stronger generalizability than individual models in many fields of machine learning \cite{zhou2012ensemble}, we create $n$ ensemble models $\{g_i\}_{i=1}^n$ by randomly combining a subset of $s$ models out of $\{h_j\}_{j=1}^{m}$:
	 \begin{align}
	  		g_i(x) = & \frac{1}{m} \sum_{j=1}^{m} \alpha_{ij} h_j(x),
	  		\quad i = 1,2,\ldots,n,
	  		\label{eq:ensemble}
	 \end{align}
    where $\alpha_{ij}=1$ if $h_j$ is selected to create $g_i$, and $\sum_{j} \alpha_{ij}=s$. In Eq. \eqref{eq:ensemble}, $\{h_j\}_{j=1}^n$ have been mapped to the same perceptual scale such that the weighted summation is legitimate.
    
\subsection{Failure Identification}
\label{sect:falilure_identification}
Given the large-scale unlabeled dataset $\mathcal{S}$, gMAD selects top-$k$ pairs of images that best discriminate between the target model $f$ and the competing model $g_i$:
\begin{align}\label{eq:gmad1}
(\hat{x}_{ik}, \hat{y}_{ik}) = &\mathop{\arg\max}_{x,y} (g_i(x) - g_i(y)) - (f(x) - f(y))\nonumber\\
    	    &\text{s.t. } x,y \in \mathcal{S}\setminus\{\hat{x}_{ij}, \hat{y}_{ij}\}_{j=1}^{k-1},
\end{align}
where $\{\hat{x}_{ij}, \hat{y}_{ij}\}_{j=1}^{k-1}$ is the set of $k-1$ pairs of images that have been selected. The roles of $f$ and $g_i$ may be switched by replacing $\mathop{\arg\max}$ with  $\mathop{\arg\min}$.
However, recursive optimization of Eq. \eqref{eq:gmad1} may simply expose different instantiations of failures with the same underlying root causes (as shown in the first row of Figure~\ref{fig:deverse}). To encourage spotting diverse failures, a fine-grained version of gMAD can be formulated as 
\begin{align}\label{eq:gmad2}
&(\hat{x}_{ik}, \hat{y}_{ik}) = \mathop{\arg\max}_{x,y} (g_i(x) - g_i(y)) - (f(x) - f(y))\nonumber\\
    	    &\text{s.t. } f(x), f(y) \in [a, b]\text{, and } x,y \in \mathcal{S}\setminus\{\hat{x}_{ij}, \hat{y}_{ij}\}_{j=1}^{k-1},
\end{align}
where $[a,b]$ defines a quality level, within which $f$ predicts the two images $x$ and $y$ to have similar quality (see the second row of Figure \ref{fig:deverse}). In this case, $f$ is regarded as the \textit{defender}, while $g_i$ is the \textit{attacker}. We may select several (non-overlapping) quality levels to cover the full quality spectrum. All image pairs selected by exhausting the competing models and the quality levels form the gMAD set $\mathcal{M}$, whose size is considerably smaller than the unlabeled set $\mathcal{S}$ and is adjustable to fit the available human labelling budget. In practice, it is possible to further diversify the spotted failures in $\mathcal{M}$ by decreasing $k$ and increasing $n$ provided that the created ensemble models differ to certain degrees (see Figure~\ref{fig:deverse}). Finally, we conduct subjective testing to collect human scores for each $(x,y)\in \mathcal{M}$, leading to four possible results:
\begin{itemize}
    \item \textbf{Case I}. Both $f$ and $g_i$ are consistent with humans in ranking the perceived quality of $x$ and $y$. This happens because $f$ is a top-performing model, while $g_i$ closely resembles $f$ as its pruned version. We may reduce the possibility of this outcome by increasing the size of $\mathcal{S}$ and selecting more suitable network pruning algorithms.
    \item \textbf{Case II}. $g_i$ is consistent with human perception, but $f$ is not. In this case, the selected $(x,y)$ constitutes a failure of $f$, and is informative for subsequent model rectification.
    \item \textbf{Case III}. $f$ is consistent with human perception, but $g_i$ is not. In this case, $f$ successfully spots a counterexample of $g_i$, which seems to deviate from the original goal. However, it is worth noting that $(x,y)$ is still useful in improving the performance of $f$ because we are co-evolving $f$ and $g_i$ in the subsequent stage of model rectification. That is, $g_i$ is also given the opportunity to learn from its failures, increasing the possibility of exposing $f$'s weaknesses in the next round  of the gMAD competition.
    \item \textbf{Case IV}. Neither $f$ nor $g_i$ is consistent with human perception. In this case, $(x,y)$ manifests itself as a double-failure result, which is the most informative in improving the generalizability of $f$ \cite{hooker2020compressed}.
\end{itemize}



\subsection{Model Rectification}
The labeled gMAD set $\mathcal{L}$ exposes aspects of weaknesses of $f$, and thus is useful for improving its generalization to the real world. To avoid  catastrophic forgetting \cite{mccloskey1989catastrophic}, we choose to combine $\mathcal{L}$ and previously trained dataset $\mathcal{D}$, and jointly fine-tune $f$ and $\{g_i\}_{i=1}^n$ on the combined set. By doing so, all models are able to learn from their respective failures and improve their generalizability for the next round of the gMAD competition. We may iterate this procedure of failure identification and model rectification  several rounds, leading to a progressive human-in-the-loop troubleshooting method for BIQA models in the wild. We denote the target and competing models in the first round as $f^{(0)}$ and $\{g_i^{(0)}\}_{i=1}^n$ respectively. In the $t$-th round, we fine-tune $f^{(t-1)}$ and $\{g_i^{(t-1)}\}_{i=1}^n$ on the combination of $\mathcal{D}$ and $\mathcal{L} = \bigcup_{j=1}^{t}\mathcal{L}^{(j)}$.
We summarize the proposed method in Algorithm \ref{al:algorithm1}.
   
\begin{algorithm}[t]
   \SetAlgoLined
   \KwIn{A training set $\mathcal{D}$, a large-scale unlabeled image set $\mathcal{S}$, a target BIQA model $f^{(0)}$, the number $r$ of fine-tuning rounds}
   \KwOut{A troubleshot BIQA model $f^{(r)}$}
   Train/fine-tune $f^{(0)}$ on $\mathcal{D}$ \\
   Prune $f^{(0)}$ to generate $\{h_j^{(0)}\}_{j=1}^{m}$ \\
   \For{$j \gets 1$ \KwTo $m$}
   {
   	Fine-tune $h_j^{(0)}$ on  $\mathcal{D}$ to recover the performance
   } 
   $\mathcal{L} \gets \emptyset$ \\
   \For{$t \gets 0$ \KwTo $r-1$}
   {   
   Randomly ensemble subsets of $\{h_j^{(t)}\}_{j=1}^{m}$ to construct $\{g_i^{(t)}\}_{i=1}^n$\\
   Compute the responses of $f^{(t)}$ on $\mathcal{S}$ \\
   $\mathcal{M}^{(t+1)} \gets \emptyset$, $\mathcal{L}^{(t+1)} \gets \emptyset$ \\
   \For{$i \gets 1$ \KwTo $n$}
   {
   	Compute the responses of $g_i^{(t)}$ on $\mathcal{S}$ \\
   	Seek pairs of images associated with $f^{(t)}$ and $g_i^{(t)}$ by solving Eq. (\ref{eq:gmad2}), and include them in $\mathcal{M}^{(t+1)}$\\
   }
   Collect  human scores for  $\mathcal{M}^{(t+1)}$ to form $\mathcal{L}^{(t+1)}$\\
   $\mathcal{L} \gets \mathcal{L} \bigcup \mathcal{L}^{(t+1)}$ \\
   Fine-tune $f^{(t)}$ on $\mathcal{D}\bigcup\mathcal{L}$ \\
   \For{$j \gets 1$ \KwTo $m$}
   {
   	Fine-tune $h_j^{(t)}$ on $ \mathcal{D}\bigcup\mathcal{L}$ \\
   } 
   }   	
   \caption{Troubleshooting BIQA models in the wild}
   \label{al:algorithm1}
\end{algorithm}

\section{Experiments}
In this section, we first describe the real experimental setups, and then provide quantitative and qualitative results to validate the feasibility of the proposed method, followed by an ablation study to test the failure-spotting efficiency of our method.
\subsection{Experimental Setups}
\paragraph{Target Model $f$} We use UNIQUE \cite{zhang2020uncertainty}, a state-of-the-art BIQA model with so far the best cross-distortion-scenario performance to our best knowledge. We retrain it on six IQA datasets, \ie, LIVE \cite{sheikh2006statistical}, CSIQ \cite{larson2010most}, KADID-10k \cite{lin2019kadid}, BID \cite{ciancio2010no}, LIVE Challenge \cite{ghadiyaram2015massive}, and KonIQ-10k \cite{hosu2020koniq}. We leave $20\%$ images for monitoring the performance changes of $f$ during troubleshooting.

\paragraph{Unlabeled Dataset $\mathcal{S}$}
   To construct the large-scale dataset $\mathcal{S}$ for gMAD to seek potential failures of $f$, we first download $750,000$ images from the Internet followed by automatic pre-screening to remove duplicate and non-photographic images. Afterward, we sample $100,000$ images with marginal distributions nearly uniform with respect to image attributes, including bitrate, JPEG compression ratio, brightness, colorfulness, contrast, and sharpness~\cite{vonikakis2016shaping}. Finally, we down-sample the images such that the long edge has $1,024$ pixels as a way of facilitating computational prediction. The constructed $\mathcal{S}$ includes a wide range of realistic camera distortions, such as sensor noise contamination, motion and out-of-focus blurring, under- and over-exposure, contrast reduction, color cast, and a mixture of these.
    
\paragraph{Competing Models $\{g_i\}_{i=1}^{n}$}   
To encourage the diversity of the competing model pool, we adopt six network pruning algorithms: oneshot magnitude pruning (OMP) \cite{han2015learning}, $\ell_1$-filter pruning \cite{li2016pruning}, $\ell_2$-filter pruning \cite{li2016pruning}, TaylorFOWeight pruning \cite{molchanov2019importance}, network slimming \cite{liu2017learning}, and FPGM pruning \cite{he2019filter}, among which OMP is unstructured weight pruning, and the others are filter pruning. Fine-tuning is conducted after each model pruning method to recover the model performance. 
In addition, we use three different pruning ratios for each method, resulting in a total of $m=18$ pruned models $\{h_j\}_{j=1}^{18}$ from $f$. We randomly combine $s=8$ out of $m=18$ pruned models, giving rises to $n=120$ ensemble models $\{g_i\}_{i=1}^{120}$. Note that ensembling requires all pruned models to use the same perceptual scale. To achieve this, we map all model predictions onto the MOS scale $[0, 100]$ of the LIVE Challenge Database \cite{ghadiyaram2015massive}, with higher values indicating better perceptual quality. The fitted mapping function can be treated as part of the pruned model. As formulated in Eq. \eqref{eq:ensemble}, ensembling is implemented by simple averaging, which gives all pruned models equal weights.

     
\begin{figure}[t]
	\begin{center}
		\includegraphics[scale=0.205]{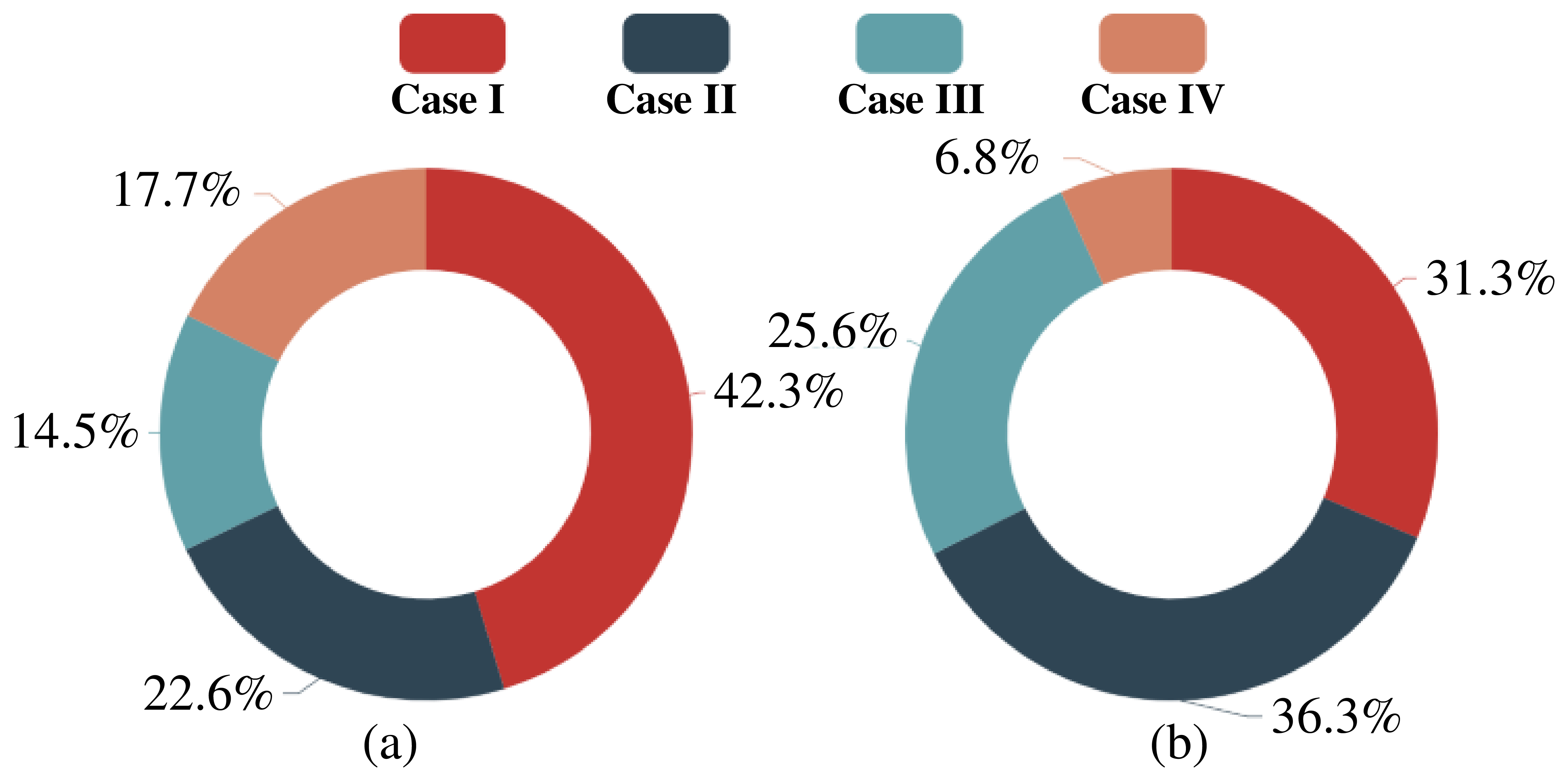}
	\end{center}
	\vspace{-.5cm}
	\caption{The empirical distributions of the gMAD pairs in  \textbf{(a)} $\mathcal{L}^{(1)}$ and \textbf{(b)} $\mathcal{L}^{(2)}$, respectively.}
	\label{fig:distribution}
\end{figure}

\paragraph{Labeled gMAD Set $\mathcal{L}$} 
To seek gMAD pairs, we set five quality levels, roughly covering the full quality range from ``bad'', ``poor'', ``fair'', ``good'', to ``excellent''. Two types of pairs are queried by treating the target model $f$ in Eq. \eqref{eq:gmad2} as the defender and the attacker, respectively. We retain a single pair that best differentiates between $f$ and $g_i$ at each quality level by setting $k=1$. We perform two rounds of troubleshooting (\ie, $r=2$), and the number of gMAD pairs in $\mathcal{L}^{(1)}$ and $\mathcal{L}^{(2)}$ are $1,184$ and $1,194$, respectively.
We gather human data from $20$ subjects in an office environment with a calibrated display~\cite{video2000final} using the single stimulus continuous quality rating. We process the raw subjective data using the outlier detection and subject rejection algorithm in~\cite{bt2002methodology}. We find that all subjects are valid, and  $2.89\%$ and $3.03\%$ of ratings are outliers and subsequently removed. 
 
%

 \begin{table}[t]
    \caption{ Correlation between model predictions and MOSs on the test set of KonIQ-10k~\cite{hosu2020koniq}, $\mathcal{L}^{(1)}$ and $\mathcal{L}^{(2)}$, respectively. Top section lists two knowledge-driven models. Middle section contains three DNN-based models retrained on the training set of KonIQ-10k. Note that the result of $f$ on $\mathcal{L}^{(2)}$ is obtained by fine-tuning it on both $\mathcal{D}$ and $\mathcal{L}^{(1)}$.}
    \label{table:comparision}
    \vspace{-.3cm}
	\begin{center}
		\begin{tabular}{l|ccc}
    		\toprule[1pt]
			SRCC  & KonIQ-10k & $\mathcal{L}^{(1)}$ & $\mathcal{L}^{(2)}$ \\
			\hline
			NIQE \cite{mittal2012making} & $0.521$ & $0.340$ & $0.293$ \\
			HOSA \cite{xu2016blind} & $0.520$ & $0.336$  & $0.287$\\
			\hline
			DB-CNN \cite{zhang2018blind}& $0.806$ & $0.690$ & $0.641$ \\
			MetaIQA \cite{zhu2020metaiqa} & $0.841$ & $0.801$ & $0.751$ \\
			HyperIQA \cite{su2020blindly} & $0.902$ & $0.802$ & $0.765$ \\
			\hline
			UNIQUE \cite{zhang2020uncertainty} (as $f$) & $0.862$ & $0.570$ & $0.553$ \\
			\hline
			\hline
			PLCC & KonIQ-10k & $\mathcal{L}^{(1)}$ & $\mathcal{L}^{(2)}$  \\
		    \hline
			NIQE & $0.529$ & $0.339$ & $0.293$ \\
			HOSA &$0.519$ & $0.327$  & $0.275$\\
			\hline
			DB-CNN &$0.828$ & $0.722$ & $0.639$ \\
			MetaIQA &$0.878$ & $0.799$ & $0.742$ \\
			HyperIQA & $0.921$ & $0.807$ & $0.741$ \\
			\hline
			UNIQUE (as $f$) & $0.875$\tnote{2} & $0.565$ & $0.555$ \\
			\bottomrule[1pt]
		\end{tabular}
	\end{center}
\end{table}

\paragraph{Fine-Tuning Details}
  For each round of model rectification, we fine-tune all $19$ models, including the target model with the same optimization settings. Specifically, the Adam method is used with a learning rate of $10^{-5}$ and a mini-batch size of $32$ - half from the previous training set $\mathcal{D}$ and half from the gMAD set $\mathcal{L}$. The maximum epoch number is set to ten. During fine-tuning, we re-scale and crop the images to $384 \times 384$. We test on images of original sizes. It takes about $162$ GPU hours for each round of fine-tuning as measured on a machine with a single RTX 2080Ti.
 
\begin{figure*}[t]
	\centering
	\subfloat[]{\includegraphics[width=0.23\textwidth]{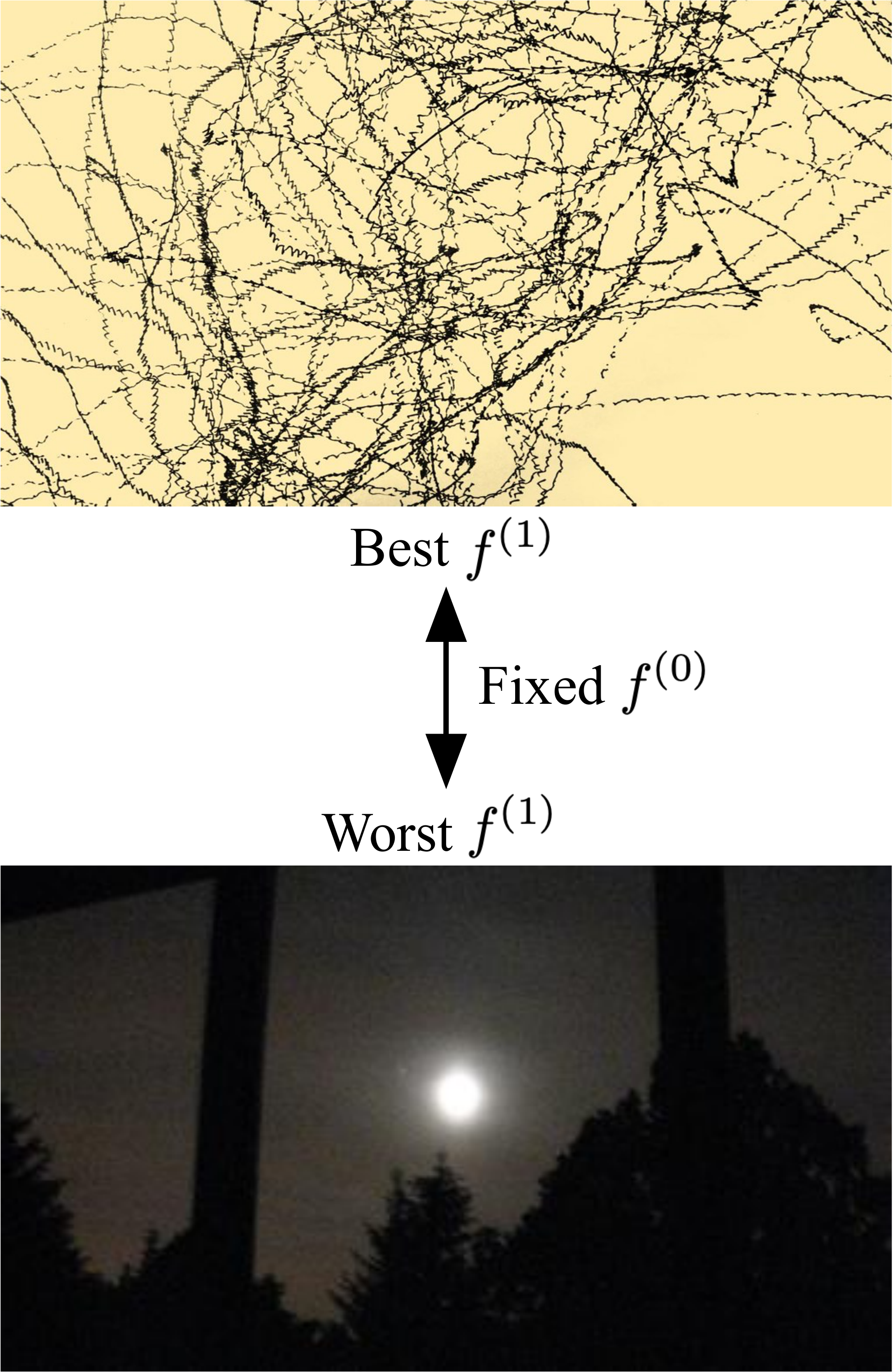}}\hskip.25em
	\subfloat[]{\includegraphics[width=0.23\textwidth]{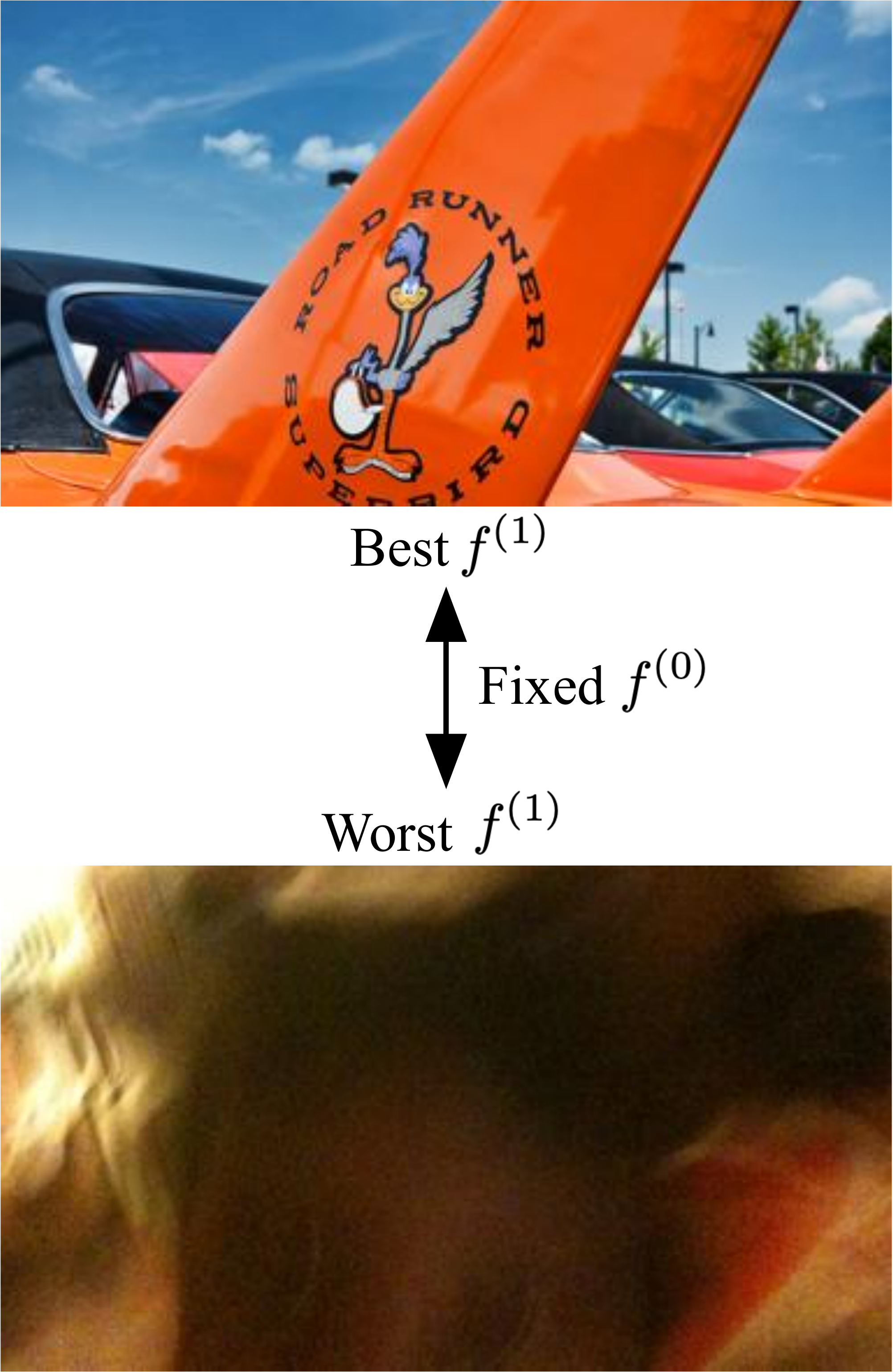}}\hskip.25em
	\subfloat[]{\includegraphics[width=0.23\textwidth]{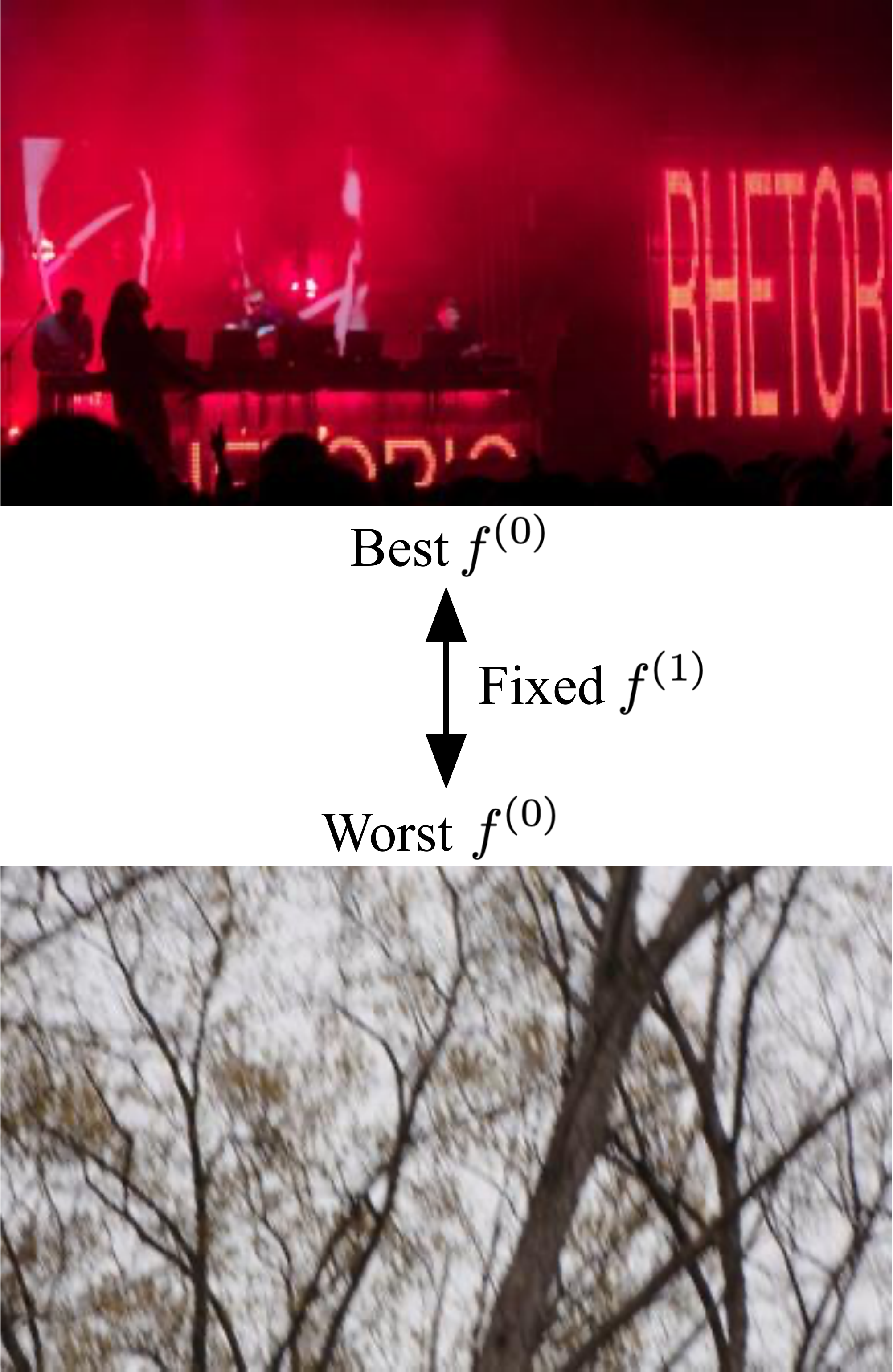}}\hskip.25em
	\subfloat[]{\includegraphics[width=0.23\textwidth]{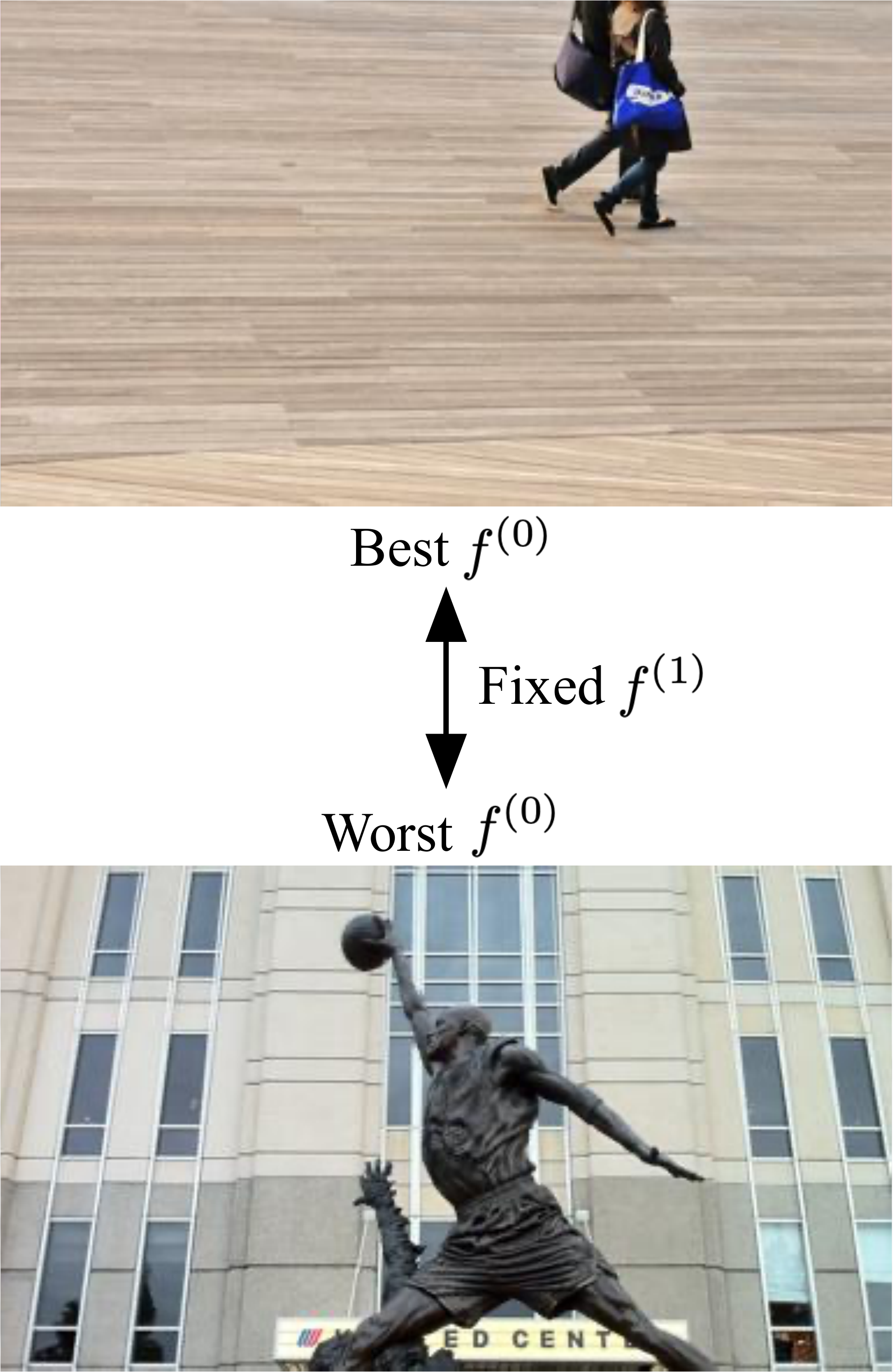}}
	\caption{Representative gMAD pairs between $f^{(0)}$ and $f^{(1)}$. \textbf{(a)} Fixing $f^{(0)}$ at the low quality level. 
\textbf{(b)}  Fixing $f^{(0)}$ at the high quality level. 
\textbf{(c)} Fixing $f^{(1)}$ at the low quality level. 
\textbf{(d)} Fixing $f^{(1)}$ at the high quality level. 
}
	\label{fig:f0_f1}
\end{figure*}

\subsection{Main Results}
\paragraph{Failure Identification} Table \ref{table:comparision} lists the Spearman rank correlation coefficient (SRCC) and Pearson linear correlation coefficient (PLCC) results between model predictions and MOSs on the gMAD sets $\mathcal{L}^{(1)}$ and $\mathcal{L}^{(2)}$. We also include the performance on the test set of KonIQ-10k \cite{hosu2020koniq} for reference. Several aspects of the results are worth noting. First, the correlation numbers of $f$ on $\mathcal{L}^{(1)}$ and $\mathcal{L}^{(2)}$ are much lower than that on KonIQ-10k, indicating the effectiveness of  our method in exposing failures of a ``top-performing'' BIQA  model. Second, despite fine-tuned on $\mathcal{L}^{(1)}$, $f^{(1)}$ delivers slightly worse performance on $\mathcal{L}^{(2)}$ compared to $f^{(0)}$ on $\mathcal{L}^{(1)}$. This suggests that the co-evolving ensemble models are able to spot stronger errors of $f$ in the second round of troubleshooting. Third, the identified counterexamples of $f$ in each round show increasing transferability to falsify five
existing BIQA models, as evidenced by larger performance drops on $\mathcal{L}^{(2)}$ than $\mathcal{L}^{(1)}$. 

In Figure \ref{fig:distribution}, we take a look at the empirical distributions of four possible results of gMAD pairs (see Section \ref{sect:falilure_identification}).  Generally, ensemble models are more aggressive to falsify the target model, and are more resistant to the target's attacks as well. After the first round of fine-tuning, the failure-spotting capability of all models has been significantly improved.




    \begin{figure*}[t]
    	\centering
    	\subfloat[]{\includegraphics[width=0.23\textwidth]{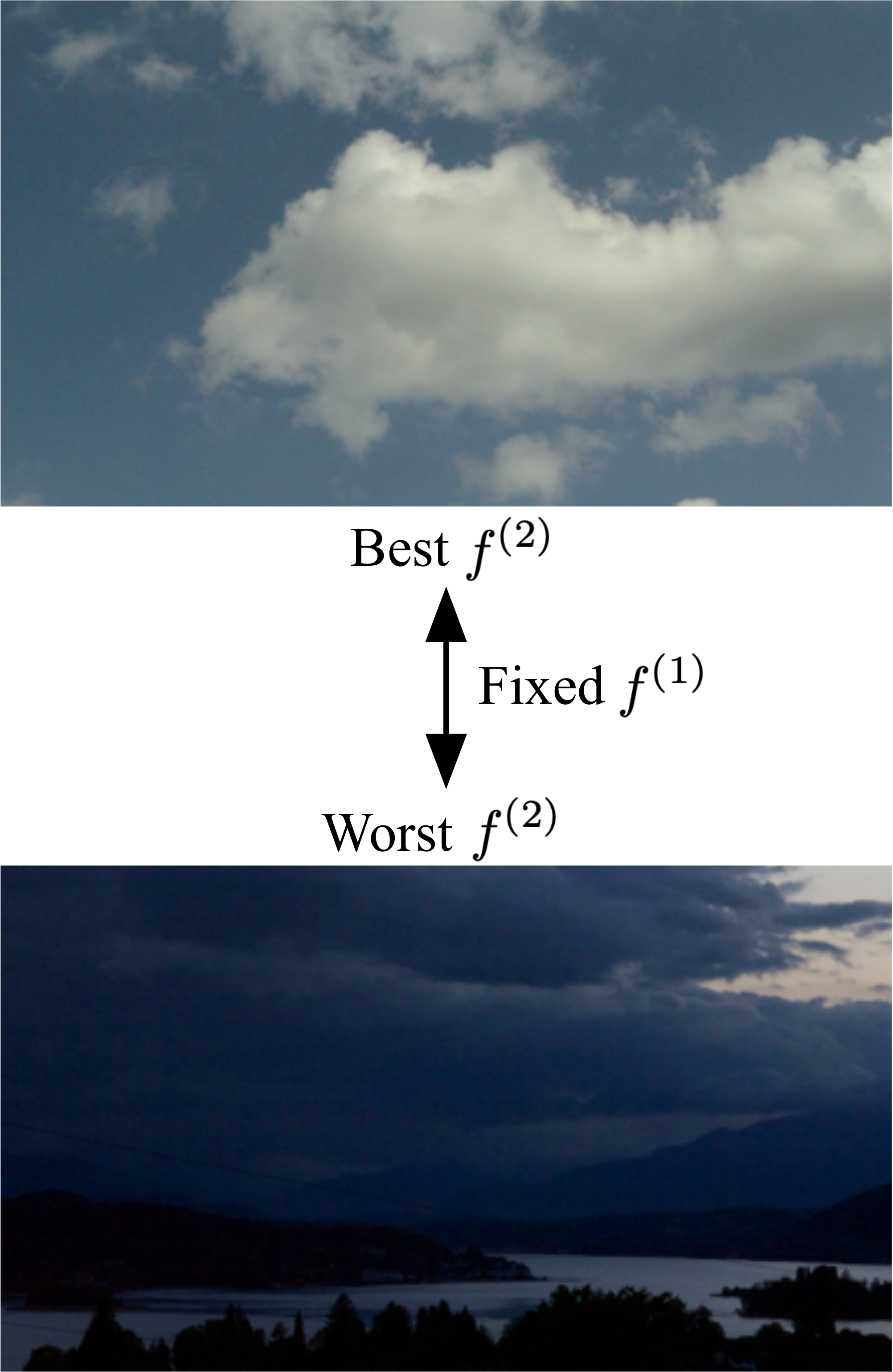}}\hskip.25em
    	\subfloat[]{\includegraphics[width=0.23\textwidth]{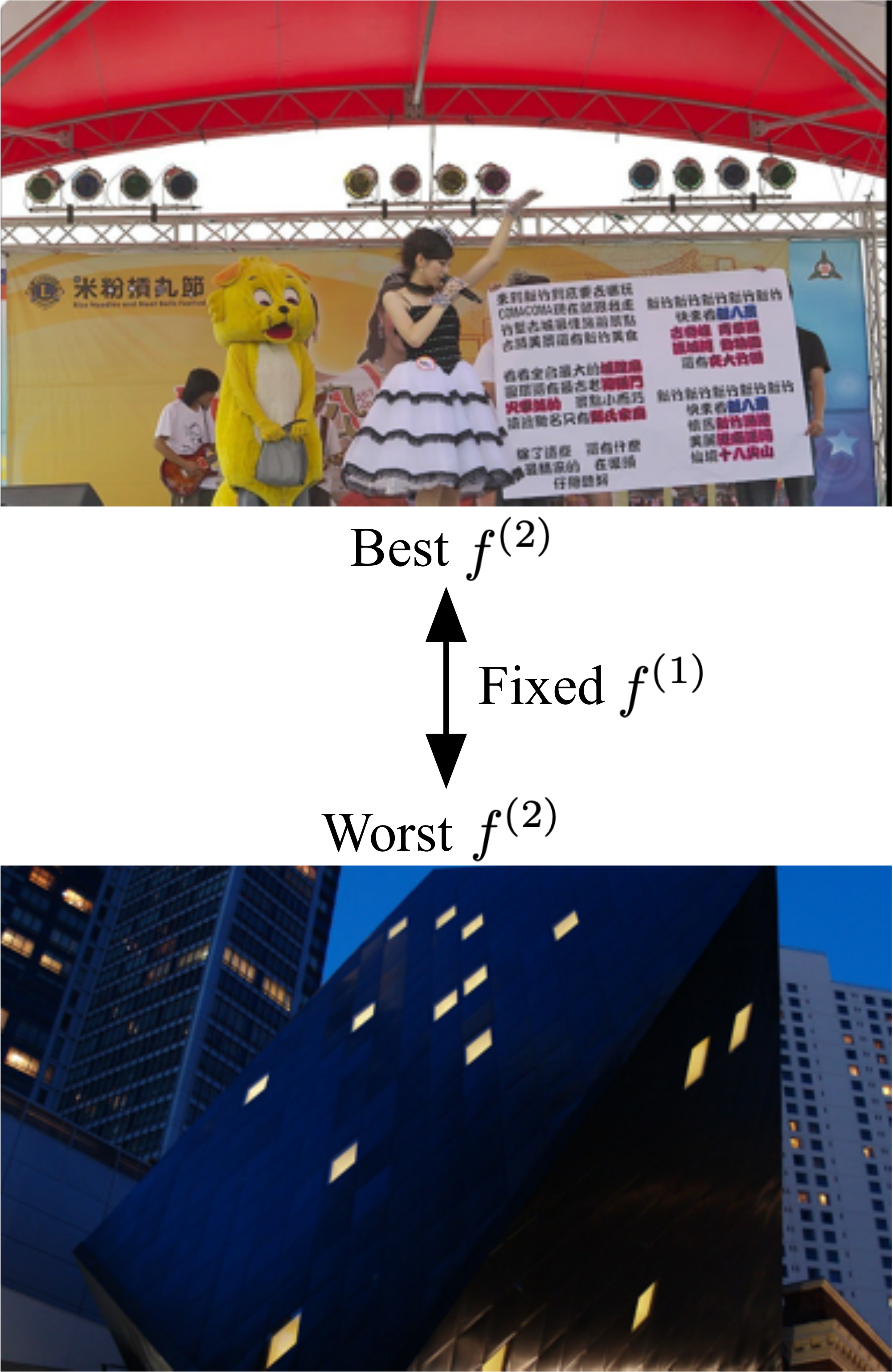}}\hskip.25em
    	\subfloat[]{\includegraphics[width=0.23\textwidth]{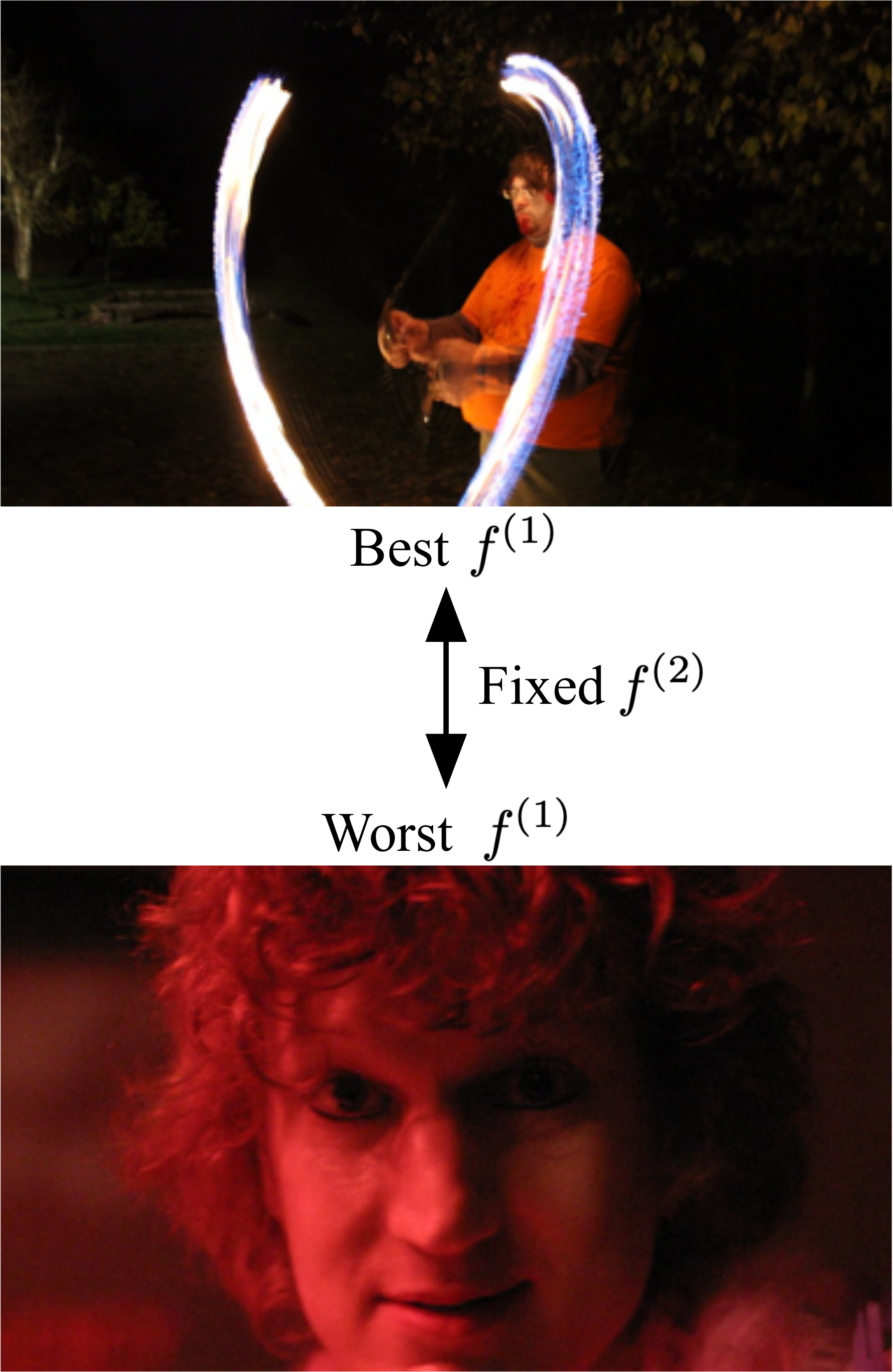}}\hskip.25em
    	\subfloat[]{\includegraphics[width=0.23\textwidth]{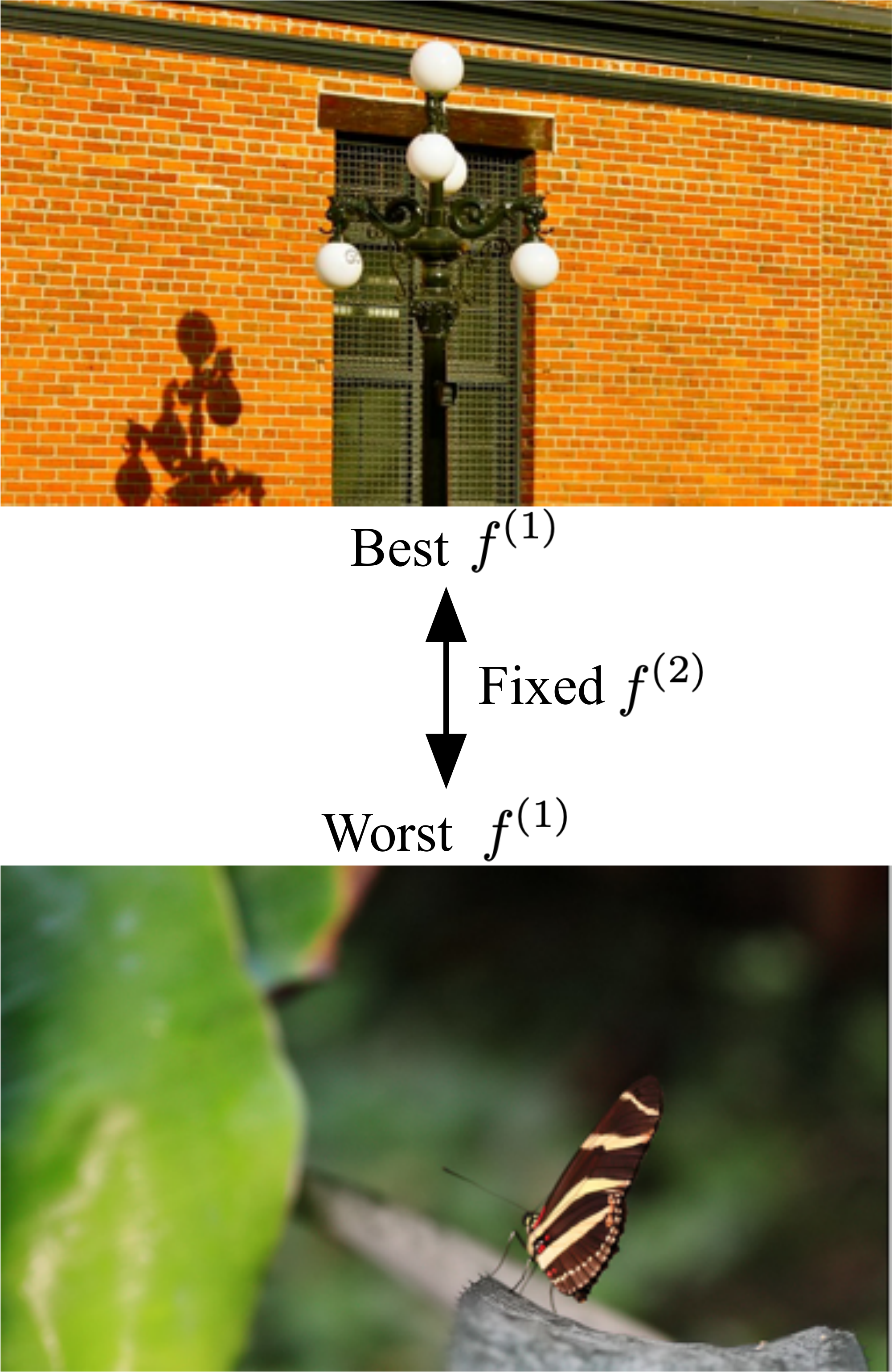}}
    	\caption{Representative gMAD pairs between $f^{(1)}$ and $f^{(2)}$. \textbf{(a)} Fixing $f^{(1)}$ at the low quality level. 
    \textbf{(b)} Fixing $f^{(1)}$ at the high quality level. 
    \textbf{(c)}  Fixing $f^{(2)}$ at the low quality level.  
    \textbf{(d)} Fixing $f^{(2)}$ at the high quality level. 
    }
    	\label{fig:f1_f2}
    \end{figure*}
    
        \begin{table}[t]
        \caption{Global ranking results of $f^{(0)}$, $f^{(1)}$ and $f^{(2)}$ in gMAD. A larger aggressiveness/resistance value indicates better performance \cite{ma2018group}.}
        \vspace{-.3cm}
    	\begin{center}
    		\begin{tabular}{c|cc}
    			\toprule[1pt]
    			& Aggressiveness & Resistance \\
    			\hline
    			$f^{(0)}$ & $-2.833$ & $-0.053$ \\
    			$f^{(1)}$ & $1.366$ & $0.019$  \\
    			$f^{(2)}$ & $1.467$ & $0.034$  \\
    			\bottomrule[1pt]
    		\end{tabular}
    	\end{center}
    	\label{table:f0f1f2}
    \end{table}

    \paragraph{Model Rectification} We progressively fine-tune the target model $f^{(0)}$ on $\mathcal{D}\bigcup\mathcal{L}^{(1)}$ to obtain $f^{(1)}$, which is further fine-tuned on  $\mathcal{D}\bigcup\mathcal{L}^{(1)}\bigcup\mathcal{L}^{(2)}$ to obtain $f^{(2)}$. To verify the relative improvements resulting from model rectification, we let $f^{(0)}$, $f^{(1)}$, and $f^{(2)}$ play the gMAD game against one other on $\mathcal{S}\setminus\mathcal{L}$. In each of $\binom{3}{1}$ competitions, we select $100$ gMAD pairs at five quality levels for human annotation. As suggested in \cite{ma2018group}, we aggregate the paired comparison results into two global ranking vectors to indicate how \textit{aggressive} one model is to falsify other models as the attacker and how \textit{resistant} one model is to survive other models' attacks as the defender. Table \ref{table:f0f1f2} 
   shows the global ranking results with higher values indicating better performance. It is easy to conclude that $f$ continually evolves to be a better model in terms of both aggressiveness and resistance in gMAD without forgetting previously seen data. This verifies the feasibility of the proposed scheme to troubleshoot BIQA models in the wild.

    \begin{figure*}[t]
    	\centering
    	\subfloat[]{\includegraphics[width=0.23\textwidth]{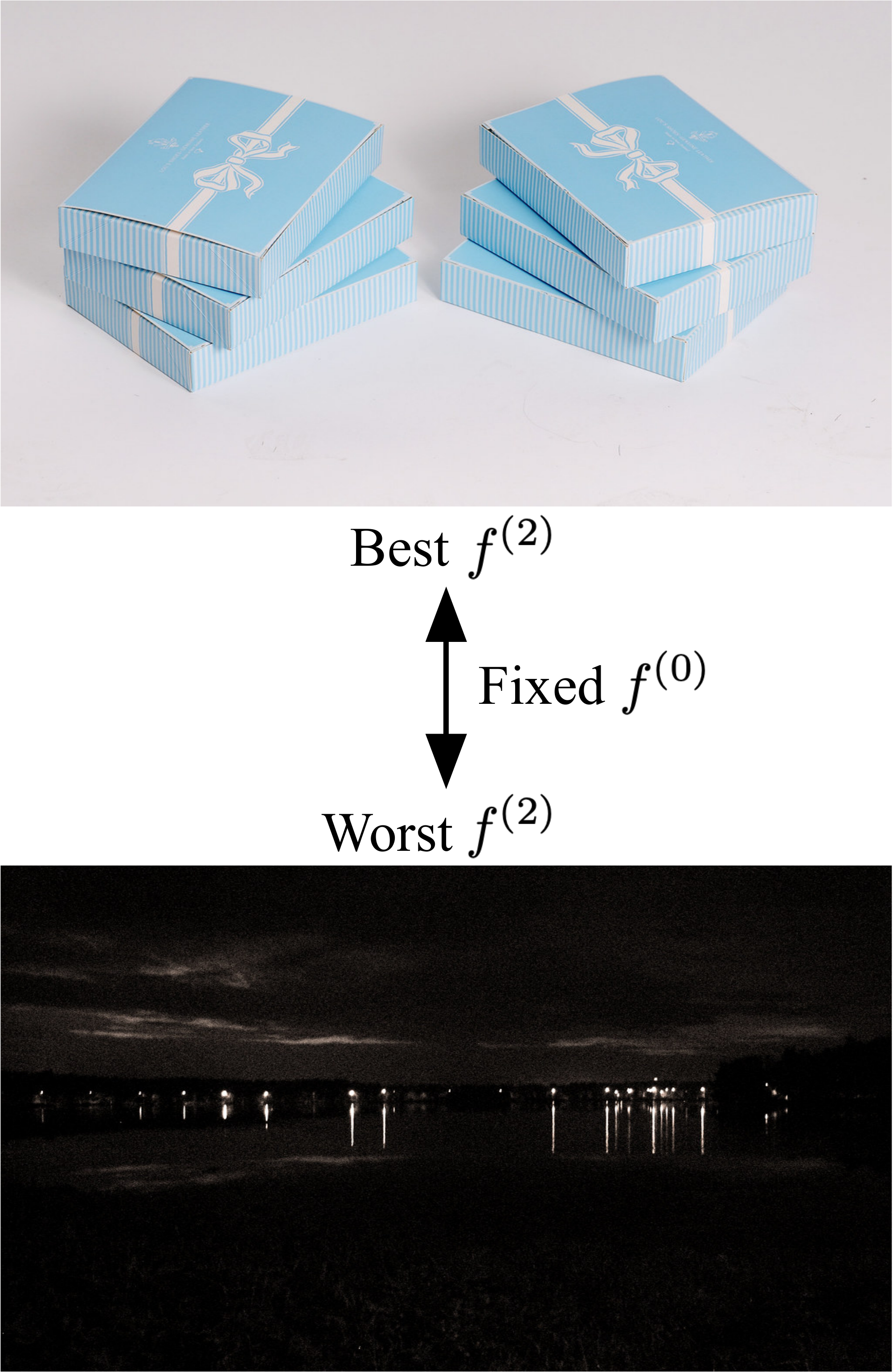}}\hskip.25em
    	\subfloat[]{\includegraphics[width=0.23\textwidth]{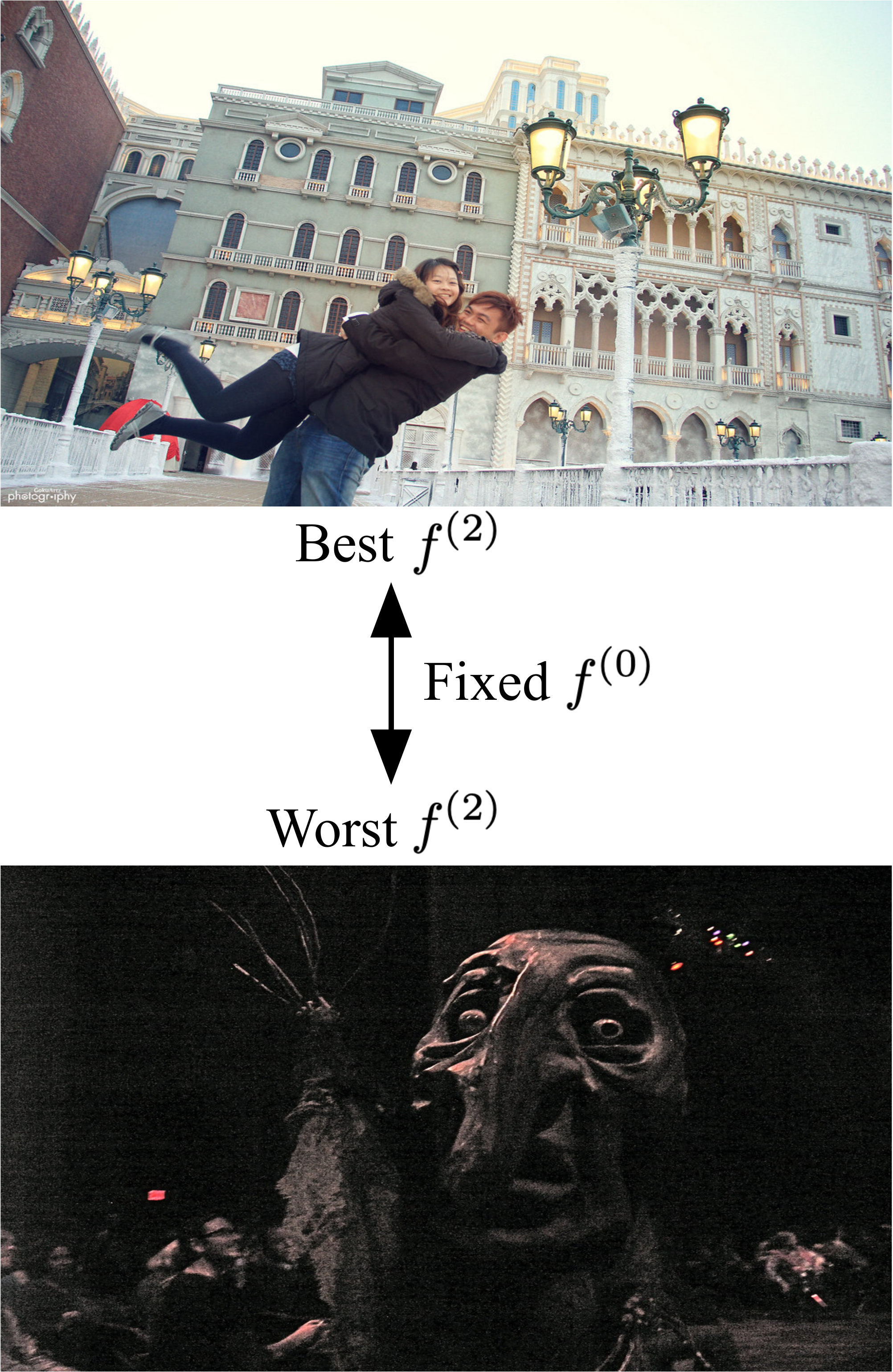}}\hskip.25em
    	\subfloat[]{\includegraphics[width=0.23\textwidth]{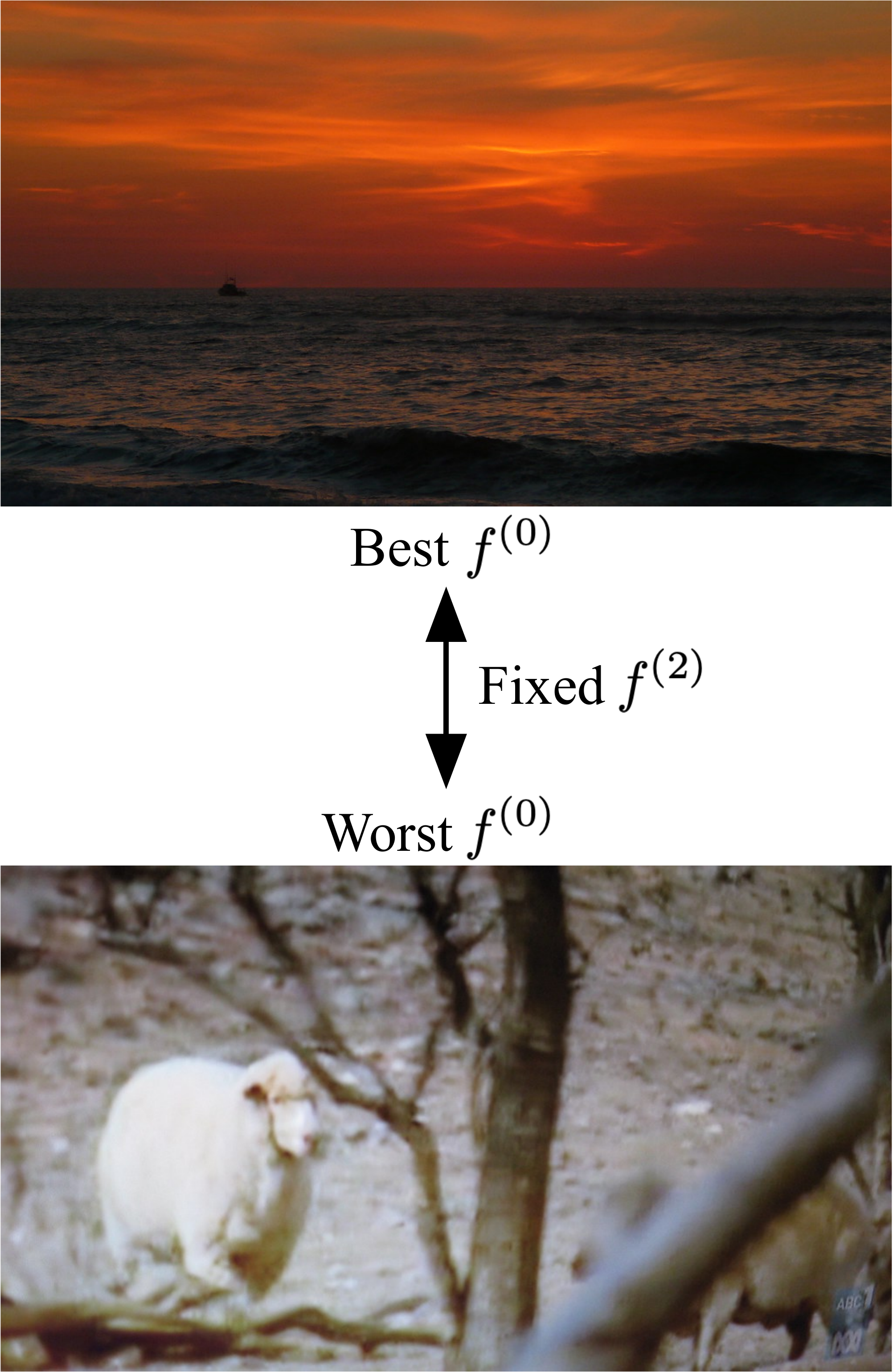}}\hskip.25em
    	\subfloat[]{\includegraphics[width=0.23\textwidth]{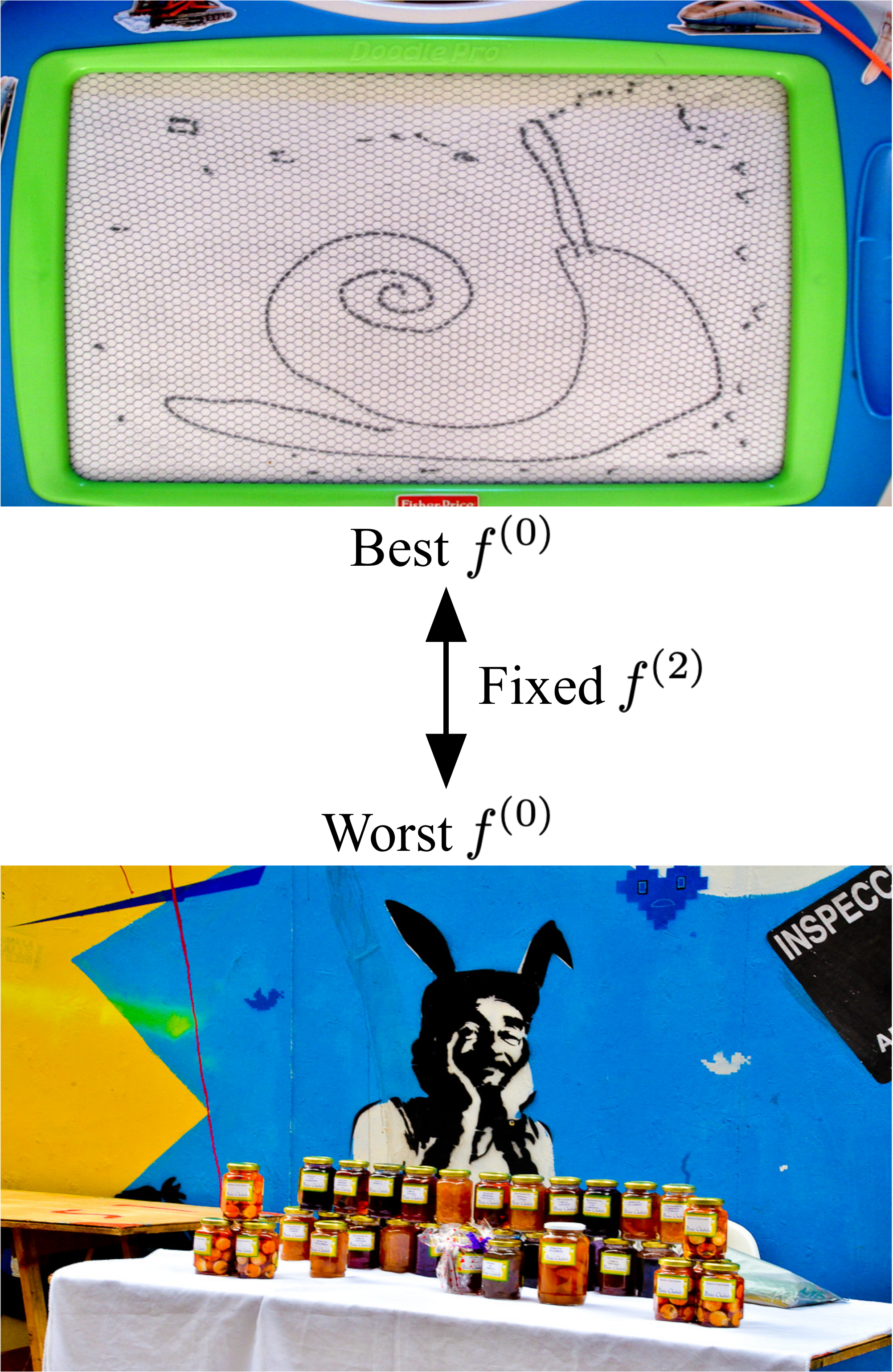}}
    	\caption{Representative gMAD pairs  between $f^{(0)}$ and $f^{(2)}$. \textbf{(a)} Fixing $f^{(0)}$ at the low quality level.
    \textbf{(b)} Fixing $f^{(0)}$ at the high quality level. 
    \textbf{(c)}  Fixing $f^{(2)}$ at the low quality level. 
    \textbf{(d)} Fixing $f^{(2)}$ at the high quality level. 
    }
    	\label{fig:f0_f2}
    \end{figure*}

    We next compare $f^{(0)}$, $f^{(1)}$, and  $f^{(2)}$ qualitatively.
    Figure \ref{fig:f0_f1} shows four representative gMAD pairs between $f^{(0)}$ and $f^{(1)}$. It is clear that the pairs of images in (a) and (b) exhibit substantially different quality, which is in disagreement with $f^{(0)}$. In contrast, $f^{(1)}$ correctly predicts top images to have much better quality than bottom images. When the roles of $f^{(0)}$ and $f^{(1)}$ are reversed, $f^{(0)}$ still fails to expose failures of $f^{(1)}$ (see (c) and (d)), suggesting that $f^{(1)}$ is significantly improved by learning from the gMAD set in the first round. 

    Figure \ref{fig:f1_f2} depicts four gMAD competition results between $f^{(1)}$ and $f^{(2)}$. In (a) and (b), we observe that $f^{(2)}$ is able to falsify  $f^{(1)}$ by finding its counterexamples which appear dark, but the perceptual gaps between best and worst cases are not as large as those when $f^{(1)}$ attacks $f^{(0)}$. In (c) and (d),  $f^{(2)}$ successfully survives the attacks from $f^{(1)}$,  with pairs of images of similar quality according to human perception. This indicates that perceptual gains from the second round of fine-tuning are not as substantial as those in the first round,  which is a common phenomenon  in active learning.


    We also show four gMAD pairs between $f^{(0)}$ and $f^{(2)}$ in Figure \ref{fig:f0_f2} to further demonstrate the improvements of $f^{(0)}$ after two rounds of troubleshooting.  $f^{(2)}$ favors the top images in (a) and (b), which is consistent with human judgments, suggesting that $f^{(2)}$ successfully attacks $f^{(0)}$. $f^{(0)}$ fails to penalize the top image in (c) and (d), which are spotted by $f^{(2)}$. This further validates the improved generalizability of $f$ to the real world.

\subsection{Ablation Study} \label{sec:compare_al}
    In this subsection, we show that gMAD sampling in our method  has stronger failure-spotting capability compared to five active learning methods for regression: random sampling, QBC \cite{seung1992query}, EMCM \cite{cai2013maximizing}, RSAL \cite{douak2011two,douak2013kernal}, and GS \cite{bhaskara2019greedy}. We conduct experiments on the smartphone photography attribute and quality (SPAQ) dataset
    \cite{fang2020cvpr}, which contains $11,125$ human-rated images captured by $66$ smartphones.
    We sample a subset of $200$ images by each method, and compute the SRCC and PLCC between MOSs and predictions by $f^{(0)}$ (\ie, UNIQUE \cite{zhang2020uncertainty}). 
    Table \ref{table:ablationstudy} shows the results, with a lower correlation coefficient indicating better performance. As can be seen, the images selected by gMAD \cite{ma2018group} lead to the worst performance of $f^{(0)}$, among all methods, which shows the failure-spotting capability of the proposed gMAD sampling.

    \begin{table}
    	\caption{Comparison of the failure-spotting efficiency  between gMAD sampling and
        five active learning methods  on the SPAQ database \cite{fang2020cvpr}. A lower correlation coefficient indicates better performance.}
        \vspace{-.3cm}
    	\begin{center}
    		\begin{tabular}{l|cc}
    			\toprule[1pt]
    			Method & SRCC  & PLCC  \\
    			\hline
    			Random & $0.729$ & $0.776$ \\
    			QBC  & $0.180$ &$0.334$ \\
    			RSAL & $0.629$ &$0.700$ \\
    			EMCM  & $0.547$ & $0.584$ \\
    			GS & $0.395$ & $0.392$ \\
    			\hline
    			Ours & $\mathbf{0.137}$ & $\mathbf{0.297}$ \\
    			\bottomrule[1pt]
    		\end{tabular}
    	\end{center}
    
    	\label{table:ablationstudy}
    \end{table}
    
\section{Conclusion}
    We have introduced a computational method for progressively troubleshooting BIQA models in the wild. The key to success of our method is to construct strong ``self-competitors'' as random ensembles of pruned versions of a ``top-performing'' target model. We have demonstrated the effectiveness of the ensemble models in exposing diverse counterexamples of the target model in the gMAD competition. A second advantage of our method is the flexibility to co-evolve the target and competing models, which allows all models to learn from their respective failures, making progressively troubleshooting the target model more effective.
    
    Our work extends a new line of research in BIQA with many important topics to be explored. For example, the current work only performs two rounds of troubleshooting due to the limited human labeling budget. Nevertheless, it is interesting to mathematically analyze the convergence of the proposed method or come up with a practical stop criterion to guide the setting of the fine-tuning round. Moreover, the computational complexity of constructing competing models, \ie, pruning followed by fine-tuning, is relatively high. It is thus worth exploring more computationally efficient methods, \eg, snapshot ensembles \cite{gao2017snapshot} for competing model construction. Another  future direction is to extend the current work to troubleshoot BIQA models in the laboratory and wild, towards universal and generalizable BIQA.

\section*{Acknowledgments}
The authors would like to thank all subjects who
participated in our subjective study during this period of the coronavirus pandemic. This work was supported in part by the National Natural Science Foundation of China (62071407), and the CityU SRG-Fd and APRC Grants (7005560 and 9610487).

{\small
\bibliographystyle{abbrv}
\bibliography{egbib}
}

\end{document}